\documentclass[11pt,a4paper]{article}

\usepackage[margin=25mm]{geometry}
\usepackage{graphicx}
\usepackage{newtxtext}
\usepackage{placeins}
\usepackage{amsmath,amssymb}
\usepackage{setspace}
\usepackage[numbers,sort&compress]{natbib}
\usepackage[hidelinks]{hyperref}
\usepackage{url}

\setcitestyle{numbers,square,sort&compress}
\begin{document}

\title{Fu-TCM: a multimodal large language model that learns traditional Chinese medicine diagnostic reasoning}

\author{\parbox{0.92\textwidth}{\centering
Jiahang Xie$^{1}$, Tong Jin$^{5}$, Lingfei Wu$^{2}$, Yaping You$^{2}$,\\
Jie Huang$^{1}$, Yuanchang Jin$^{1}$, Weikang Gong$^{1}$, Wei Cheng$^{6}$,\\
Jia You$^{3,4}$, Pan Sun$^{2,*}$, Jianfeng Feng$^{3,*}$, Hongcheng Guo$^{1,*}$\\[1em]
\small
$^1$School of Data Science, Fudan University, Shanghai, China\\
$^2$The Affiliated People's Hospital of Fujian University of Traditional Chinese Medicine, Fuzhou, Fujian, China\\
$^3$Institute of Science and Technology for Brain-Inspired Intelligence, Fudan University, Shanghai, China\\
$^4$Key Laboratory of Computational Neuroscience and Brain-Inspired Intelligence, Fudan University, Ministry of Education, Shanghai, China\\
$^5$School of Life Sciences, Beijing University of Chinese Medicine, Beijing, China\\
$^6$Institute of Science and Technology for Brain-Inspired Intelligence, Department of Neurology, Huashan Hospital, Fudan University, Shanghai, China\\
\vspace{0.75em}
$^*$Corresponding authors: fjsrmyy@gmail.com (P.S.); jianfeng64@gmail.com (J.F.); guohc@fudan.edu.cn (H.G.)
}}
\date{}

\begin{singlespace}
\maketitle
\end{singlespace}

\begin{abstract}
Large language models (LLMs) are advancing in multimodal understanding and reasoning, accelerating medical model development. Traditional Chinese medicine (TCM) uses evidence from the four examinations to determine disease location and nature, differentiate syndromes and guide treatment. This diagnostic process constitutes a complete chain of reasoning. We present Fu-TCM, a multimodal TCM LLM trained on data retained after cleaning and filtering an initial collection of approximately 7.12 million examples derived from classical texts, textbooks, medical images, clinical cases and publicly available datasets. The model was further optimized using Bianzheng-Grounded Policy Optimization (BGPO), which explicitly targets syndrome differentiation through rewards for response format, intermediate TCM reasoning fields and final syndrome predictions. Fu-TCM-27B led all six benchmarks spanning TCM reasoning, text and vision, achieving a macro-average accuracy of 81.01\% and surpassing the strongest reference model with complete results across all six benchmarks by 7.13 percentage points. Its largest margin was 11.90 points on Fu-bianzheng. Case-level analyses further showed a positive association between intermediate-field accuracy and final syndrome-selection accuracy in most evaluated models. These findings suggest that TCM syndrome differentiation may be computationally learnable and that structured reasoning supervision can improve performance across TCM tasks.
\end{abstract}


\clearpage
\section{Introduction}

Large language models (LLMs) have rapidly improved at representing knowledge, understanding language and reasoning across modalities \cite{moor2023foundation,singhal2023medpalm,singhal2025expertmedqa,yin2024mllm,tu2023generalist,rajpurkar2022ai,acosta2022multimodal,li2023llavamed}. This progress does not mean that general-purpose LLMs can be transferred directly into clinical use. Medical decisions leave little room for error. An answer that is wrong, lacks evidence or fails to account for a patient's condition may mislead clinicians and lead to inappropriate diagnosis or treatment \cite{omiye2024potentials,kim2025inflexible}. Medical LLMs therefore require training and evaluation within specific specialties \cite{wu2025bridge,zhang2023huatuogpt,bao2023discmedllm}. Their answers should be grounded in clinical evidence, and clinicians should be able to review both the basis for a decision and the diagnostic process \cite{who2024lmm,fda2026cds}.

Traditional Chinese medicine (TCM) has its own body of knowledge and a diagnostic framework centred on TCM reasoning. Before arriving at a syndrome diagnosis, practitioners synthesize findings from the four examinations: inspection, auscultation and olfaction, inquiry, and palpation. These findings indicate disease location and nature, zang-fu involvement, and the state of qi, blood and body fluids \cite{who2022tcmterms,gbt15657}. The diagnosis then guides treatment. A TCM model should likewise acquire specialized TCM knowledge, integrate multimodal evidence from the four examinations and learn the path from clinical findings through intermediate TCM reasoning fields to the final syndrome prediction.

Existing research on TCM LLMs has concentrated on domain knowledge, adaptation to TCM tasks and multimodal evidence \cite{han2026tcmreview,li2025tcmbest4sdt,kong2025mtcmb,huang2025tcm3ceval}. Continued pretraining, instruction tuning and expert feedback have enabled general models to handle TCM question answering, consultation, prescription recommendation and case reasoning \cite{dai2024tcmchat,yang2024zhongjing,tian2024tcmllmpr,wang2026gastrotcm,wei2025biancang,hua2024lingdan,yang2024tcmgpt}. Knowledge graphs and retrieval-augmented systems provide structured access to TCM concepts and source materials \cite{zhang2024tcmkg,he2025opentcm,lewis2020rag}. Multimodal models also incorporate diagnostic images, including tongue and facial images \cite{chen2025shizhen,an2026multimodaltcmreview}. These approaches have improved the use of domain knowledge, performance on TCM tasks and integration of multimodal information. Most existing models, however, still do not explicitly model synthesis across the four examinations or TCM reasoning. They rely mainly on question and answer pairs that map inputs directly to outputs. A correct syndrome prediction therefore does not ensure that the model followed TCM reasoning \cite{turpin2023unfaithful}. Clinicians may still be unable to understand or verify how the model reached its prediction.

We developed Fu-TCM to address this gap. The multimodal framework makes the path from evidence obtained through the four examinations to intermediate TCM reasoning fields and final syndrome prediction an explicit training target. Training proceeds in two stages. TCM domain-specific learning first establishes TCM knowledge, textual understanding, visual perception and structured TCM reasoning. Bianzheng-Grounded Policy Optimization (BGPO) then assigns verifiable scores to response format, intermediate TCM reasoning fields and final syndrome predictions. Together, the two stages train the model to integrate multimodal evidence and produce a structured record from clinical findings to the final syndrome. Fu-TCM-27B ranked first on all six benchmarks covering TCM reasoning, text and vision, with its largest margin observed on Fu-bianzheng. Fu-TCM achieved the strongest overall performance among the evaluated models, showing that the framework improves not only syndrome-differentiation reasoning but also textual and visual capabilities. Besides, Fu-TCM follows the TCM diagnostic reasoning process. Clinicians can therefore understand, review and trace the complete path from evidence obtained through the four examinations to final syndrome prediction.

\section{Results}

\subsection{Framework overview}

\begin{figure}[!htbp]
\centering
\includegraphics[width=\textwidth,height=0.73\textheight,keepaspectratio]{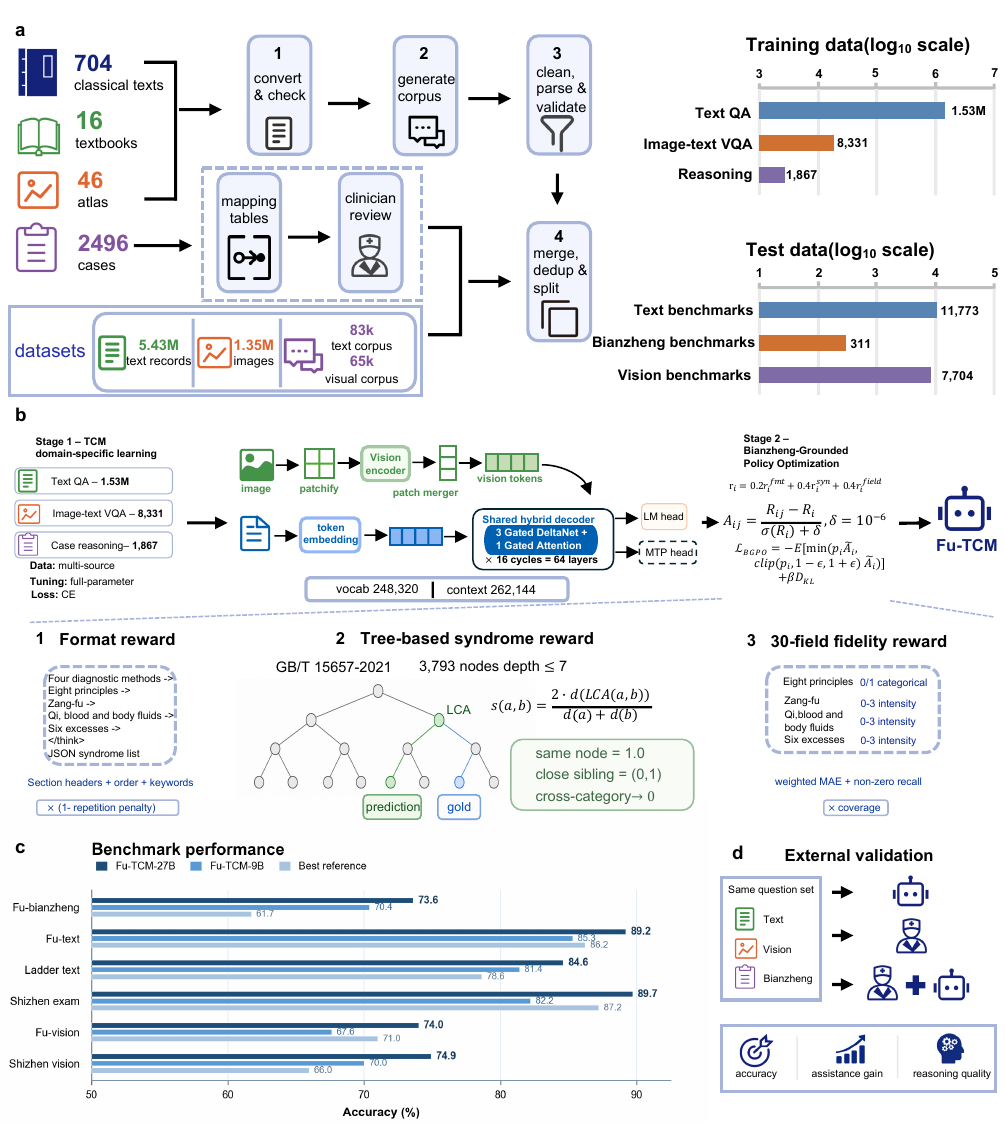}
\caption{Fu-TCM framework. a, Processing of multisource traditional Chinese medicine texts, images and clinical cases to construct text-based and image--text question--answer pairs, case-reasoning data and test datasets. b, Two-stage training with TCM domain-specific learning and Bianzheng-Grounded Policy Optimization, using rewards for response format, syndrome prediction and intermediate bianzheng fields to optimize structured bianzheng reasoning. c, Performance comparison between Fu-TCM and leading models across text, vision and TCM reasoning benchmarks. d, Same-question parallel external validation comparing model-only responses, responses from participants answering independently and responses from participants assisted by Fu-TCM.}
\label{fig:framework}
\end{figure}

Fu-TCM is a multimodal traditional Chinese medicine large language model framework designed to improve model capabilities across TCM tasks while producing structured outputs whose reasoning can be traced, reviewed and assessed. It integrates TCM texts, images and clinical cases from multiple sources. Training combines TCM domain-specific learning with Bianzheng-Grounded Policy Optimization and explicitly targets the path from evidence obtained through the four examinations to intermediate \textit{bianzheng} fields and syndrome labels. Figure~\ref{fig:framework} separates the framework into four modules. The data module converts classical texts, textbooks, image-rich medical books and clinical cases into text question and answer pairs, image-text question and answer pairs and structured case-reasoning data. The training module first establishes TCM knowledge, multimodal understanding and structured output capabilities through TCM domain-specific learning, and then applies Bianzheng-Grounded Policy Optimization with verifiable rewards to improve agreement between intermediate \textit{bianzheng} fields and syndrome predictions. The benchmark module combines self-built benchmarks with external datasets to compare Fu-TCM with leading models on text, vision and TCM reasoning tasks. In the external validation module, TCM physicians assess reasoning quality, safety and value for clinical decision support.

We collected a large body of classical TCM texts, together with modern textbooks, image-rich medical books, clinical cases and publicly available datasets. As shown in Fig.~\ref{fig:framework}a, the classical texts and textbooks were converted, cleaned and quality checked to generate text question and answer pairs. Diagnostic images from the illustrated medical books were matched to their explanations and verified to form image-text question and answer pairs. We processed the clinical cases with a clinician-reviewed mapping table from symptoms to \textit{bianzheng} fields. The resulting structured records linked evidence from the four examinations, intermediate \textit{bianzheng} fields and syndrome labels. The three data types supported TCM knowledge, multimodal understanding and case-level \textit{bianzheng} reasoning, respectively. The schema of intermediate \textit{bianzheng} fields explicitly modelled by Fu-TCM is provided in Supplementary Table~1.

Fu-TCM was trained in two stages: TCM domain-specific learning followed by Bianzheng-Grounded Policy Optimization. In the first stage, text question and answer pairs, image-text question and answer pairs and case-reasoning cold-start data were used for TCM domain-specific learning. Through this training, the model acquired TCM knowledge, multimodal understanding and structured \textit{bianzheng} outputs (Fig.~\ref{fig:framework}b). The second stage applied Bianzheng-Grounded Policy Optimization to the case data. Verifiable rewards for response format, syndrome prediction and intermediate \textit{bianzheng} fields optimized case-level \textit{bianzheng} reasoning. The syndrome-pattern tree used to calculate the syndrome-prediction reward is illustrated in Supplementary Figure~1. This sequence made the path from evidence obtained through the four examinations to intermediate \textit{bianzheng} fields and syndrome labels an explicit optimization target.

External validation followed a same-question parallel-group design. Fu-TCM, an independent-answer group and a Fu-TCM-assisted group completed the same TCM text, vision and TCM reasoning questions (Fig.~\ref{fig:framework}d). We compared model-only responses, responses from participants answering independently and responses from participants assisted by Fu-TCM. The evaluation covered accuracy, benefit from assistance and reasoning quality to assess external performance and clinical decision-support value.

\begin{figure}[!htbp]
\centering
\includegraphics[width=\textwidth,height=0.73\textheight,keepaspectratio]{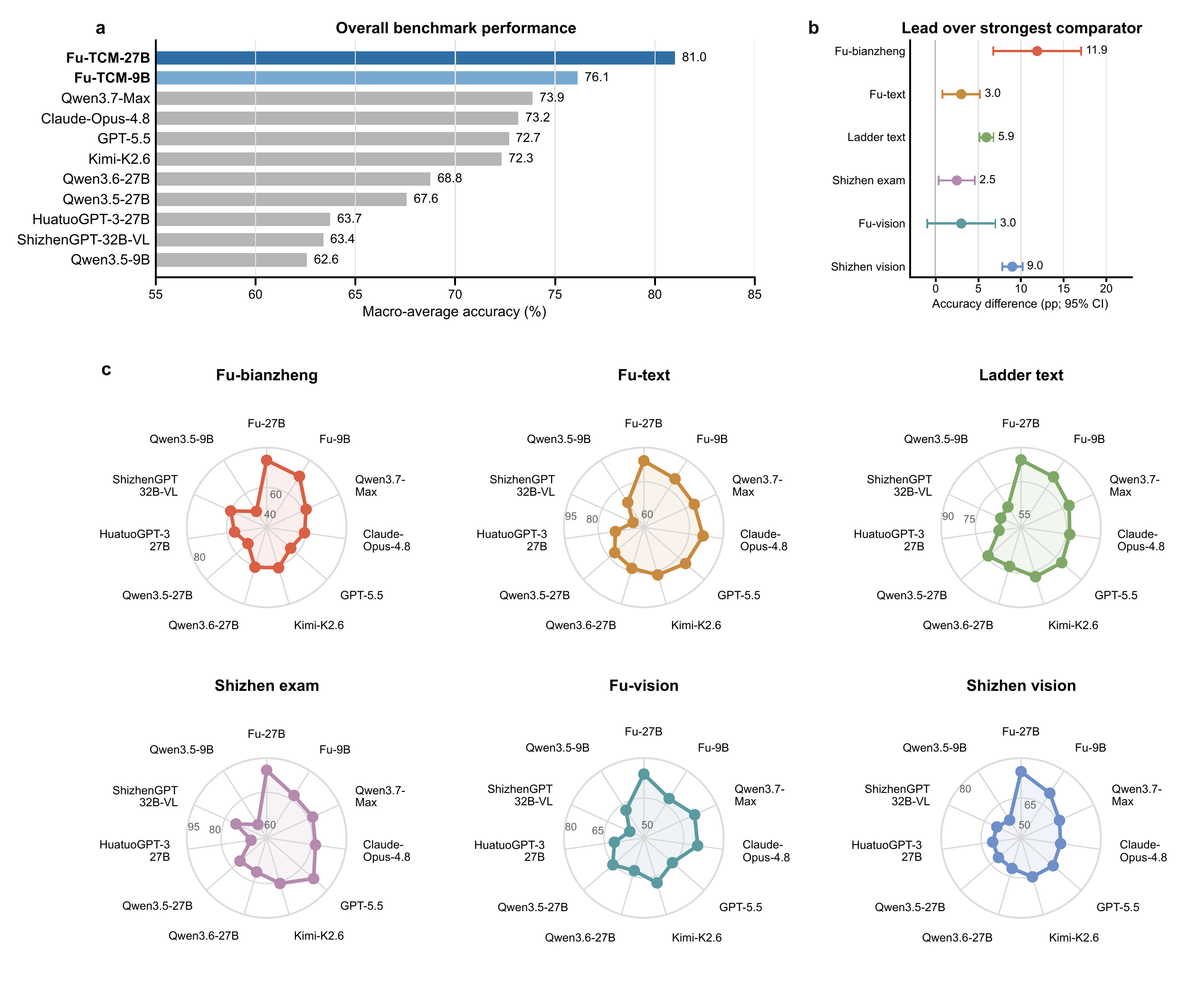}
\caption{Benchmark performance of Fu-TCM and reference models. a, Macro-average accuracy across six benchmarks for the eleven models with complete results. b, Paired accuracy differences between Fu-TCM-27B and the strongest non-Fu-TCM model on each benchmark, with matched-item 95\% bootstrap confidence intervals based on 20,000 resamples. c, Accuracy profiles across Fu-bianzheng, Fu-text, Ladder text, Shizhen exam, Fu-vision and Shizhen vision; radar axes denote models and radial distance denotes accuracy.}
\label{fig:benchmarks}
\end{figure}

\subsection{Performance across TCM benchmarks}

We used six benchmarks to evaluate Fu-TCM on TCM reasoning, TCM text knowledge and visual diagnosis (Fig.~\ref{fig:benchmarks}). Three were constructed for this study. Fu-bianzheng comprised 311 real-world clinical cases, each asking the model to select the syndrome option that best matched the clinical evidence presented in the case. Fu-text contained 1,000 questions across eight TCM knowledge categories, and Fu-vision contained 500 image-grounded questions across five diagnostic categories. The Fu-text knowledge-category distribution, Fu-vision diagnostic-category distribution and Fu-bianzheng reference-syndrome distribution are shown in Supplementary Figure~2. We also included three benchmarks from previous studies. Ladder text comprised 9,685 single- and multiple-choice questions \cite{xie2025tcmladder}. Shizhen exam comprised 1,088 questions from professional qualification and postgraduate entrance examinations \cite{chen2025shizhen}. Shizhen vision comprised 7,204 questions on medicinal recognition and visual diagnosis \cite{chen2025shizhen}.

We compared the two Fu-TCM variants with eleven reference models: Qwen3.5-9B, Qwen3.5-27B, Qwen3.6-27B, DeepSeek-V4-Pro, Kimi-K2.6, Qwen3.7-Max, GLM-5.2, GPT-5.5, Claude-Opus-4.8, HuatuoGPT-3-27B and ShizhenGPT-32B-VL \cite{chen2025shizhen,qwen2026qwen36,qwen2026qwen37max,deepseek2026v4,moonshot2026k26,glm2026glm52,openai2026gpt55,anthropic2026opus48,huatuo2026modelcard}. Accuracy was the primary evaluation metric. A single-choice response was correct when its selected option matched the reference answer. A multiple-choice response required an exact match to the full reference set. All available results were retained in benchmark-specific comparisons, but the macro-average included only models with complete results on all six benchmarks. For each benchmark, we calculated the matched-item accuracy difference between Fu-TCM-27B and the highest-scoring non-Fu-TCM model. We estimated the corresponding 95\% confidence intervals with 20,000 bootstrap resamples.

\begin{figure}[!htbp]
\centering
\includegraphics[width=\textwidth,height=0.73\textheight,keepaspectratio]{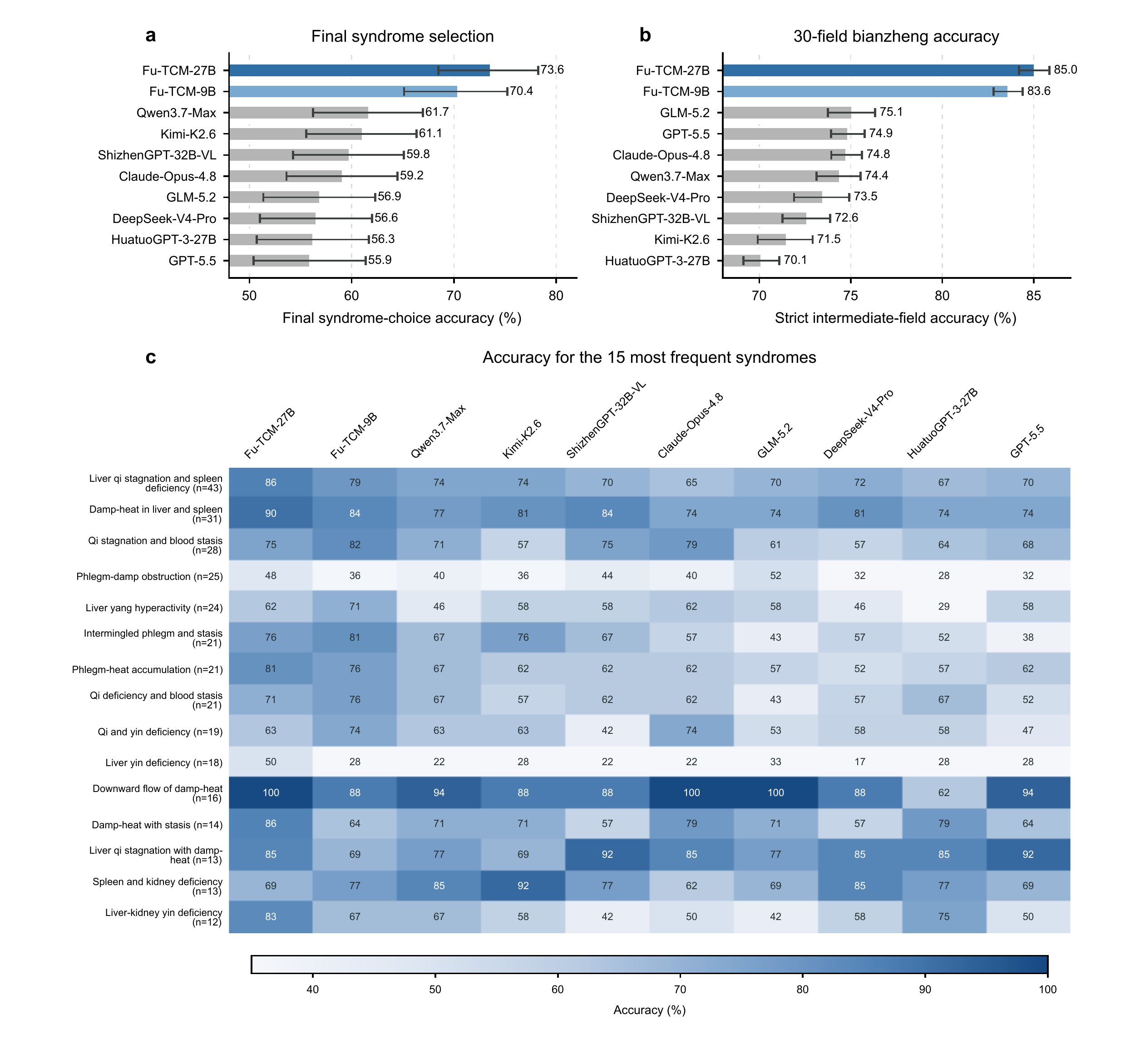}
\caption{Final syndrome-choice accuracy and intermediate bianzheng field accuracy on the Fu-bianzheng benchmark. a, Final syndrome-choice accuracy of ten models evaluated on the same 311 test cases; error bars indicate Wilson 95\% confidence intervals. b, Strict matching accuracy across 30 intermediate bianzheng fields on the same 311 cases; missing fields are scored as incorrect, and error bars indicate case-level bootstrap 95\% confidence intervals based on 4,000 resamples. c, Model accuracy for the 15 most frequent syndromes; cell values indicate accuracy (\%).}
\label{fig:bianzheng_overview}
\end{figure}

Fu-TCM-27B led all six benchmarks, with the highest macro-average accuracy of 81.01\% (Fig.~\ref{fig:benchmarks}a,c). Qwen3.7-Max was the best reference model with complete results, with a macro-average of 73.88\%, 7.13 percentage points below Fu-TCM-27B. The domain-specific models HuatuoGPT-3-27B and ShizhenGPT-32B-VL achieved macro-average accuracies of 63.74\% and 63.40\%, respectively. Compared with its Qwen3.6-27B base model, Fu-TCM-27B increased macro-average accuracy from 68.76\% to 81.01\%. Improvements on the six individual benchmarks ranged from 10.50 to 14.16 percentage points.

Paired comparisons showed where the largest performance differences occurred (Fig.~\ref{fig:benchmarks}b). Fu-TCM-27B exceeded the highest-scoring non-Fu-TCM model by 11.90 percentage points on Fu-bianzheng, 8.98 points on Shizhen vision, 5.95 points on Ladder text, 3.00 points on Fu-text and 2.48 points on Shizhen exam. The largest gaps were in visual diagnosis and TCM reasoning.

\subsection{Final syndrome and intermediate-field accuracy}

We compared ten models on Fu-bianzheng to examine final syndrome selection and the intermediate \textit{bianzheng} process. The analysis covered final syndrome-choice accuracy, strict matching accuracy across 30 intermediate \textit{bianzheng} fields and accuracy for the 15 most frequent syndromes (Fig.~\ref{fig:bianzheng_overview}). The 30 fields are diagnostic dimensions derived from established TCM differentiation systems: four eight-principle fields, five zang-fu fields, 15 fields concerning qi, blood and body fluids, and six fields for the six excesses \cite{who2022tcmterms}.

Fu-TCM-27B led both final syndrome selection and intermediate bianzheng field accuracy. Its final syndrome-choice accuracy was 73.63\% (229/311), followed by 70.42\% (219/311) for Fu-TCM-9B and 61.74\% (192/311) for Qwen3.7-Max, the strongest reference model (Fig.~\ref{fig:bianzheng_overview}a). Fu-TCM-27B exceeded Qwen3.7-Max by 11.90 percentage points (95\% paired bootstrap confidence interval, 6.75 to 17.04). The two Fu-TCM models also ranked highest in intermediate-field accuracy. Strict matching accuracy reached 85.05\% for Fu-TCM-27B (95\% case-bootstrap confidence interval, 84.20 to 85.85) and 83.61\% for Fu-TCM-9B (82.81 to 84.39). GLM-5.2, the strongest reference model on this measure, reached 75.08\% (73.74 to 76.33; Fig.~\ref{fig:bianzheng_overview}b). The two Fu-TCM models exceeded this result by 9.97 and 8.53 percentage points, respectively. The advantage was therefore present in both final syndrome selection and the intermediate bianzheng fields.

Accuracy differed substantially across syndrome categories (Fig.~\ref{fig:bianzheng_overview}c). Downward flow of damp-heat was the easiest category across the ten models, with a mean accuracy of 90.0\% and model-specific accuracies from 62.5\% to 100.0\%. Liver qi stagnation with damp-heat and damp-heat in liver and spleen were also predicted relatively accurately, with mean accuracies of 81.5\% and 79.4\%, respectively. Liver yin deficiency and phlegm-damp obstruction were consistently difficult. Their mean accuracies were 27.8\% and 38.8\%, with ranges from 16.7\% to 50.0\% and from 28.0\% to 52.0\%, respectively. Fu-TCM-27B performed best or joint best in 7 of the 15 syndrome categories. Its largest margins over the strongest syndrome-specific reference model were 16.7 percentage points for liver yin deficiency, 14.3 for phlegmheat accumulation and 11.6 for liver qi stagnation and spleen deficiency. Although liver yin deficiency was the most difficult category overall, Fu-TCM-27B achieved 50.0\% accuracy, compared with 33.3\% for the strongest reference model.

\begin{figure}[!htbp]
\centering
\includegraphics[width=\textwidth,height=0.73\textheight,keepaspectratio]{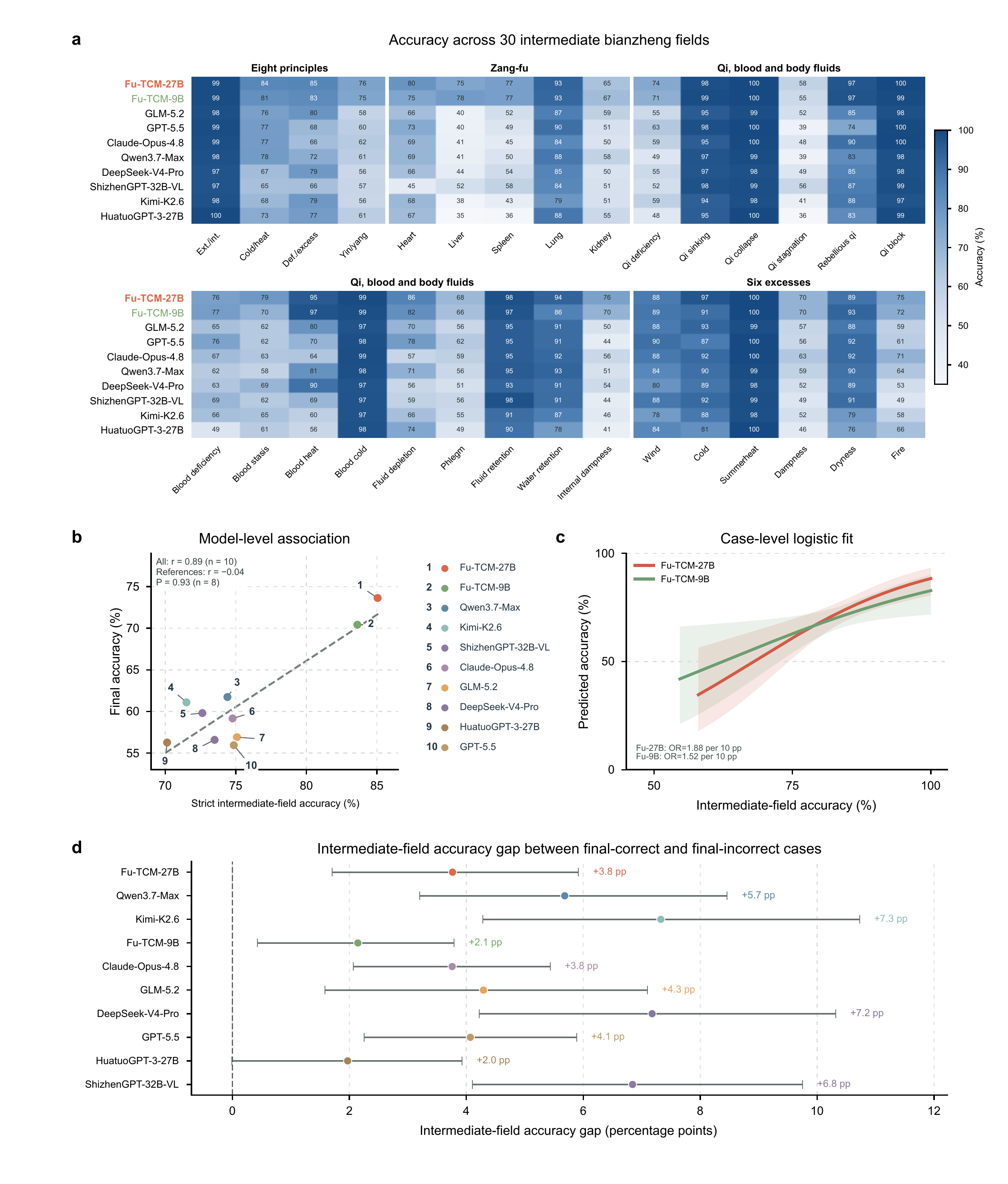}
\caption{Intermediate bianzheng field accuracy and its association with final syndrome selection on the Fu-bianzheng benchmark. a, Exact-match accuracy across 30 intermediate bianzheng fields for ten models evaluated on the same 311 cases. Each cell shows accuracy (\%). b, Association between mean intermediate-field accuracy and final syndrome-choice accuracy across ten models. Numbered points correspond to the model key. The dashed line shows the least-squares fit (Pearson r=0.89, P=0.0006; n=10 models). Within the eight reference models, the association was not statistically significant (Pearson r=$-$0.04, P=0.93). c, Case-level logistic-regression fits for Fu-TCM-27B and Fu-TCM-9B. Lines show fitted probabilities and shaded regions denote Wald 95\% confidence intervals. The odds ratios per 10-percentage-point increase in strict intermediate-field accuracy were 1.88 (95\% confidence interval, 1.35 to 2.63) and 1.52 (1.08 to 2.14), respectively. d, Difference in mean intermediate-field accuracy between cases with correct and incorrect final answers (correct minus incorrect). Points show the mean difference and error bars indicate 95\% bootstrap confidence intervals from 4,000 resamples. The intervals excluded zero for nine of the ten models.}
\label{fig:bianzheng_coupling}
\end{figure}

\subsection{Intermediate-Final Syndrome Accuracy Correlation}

We first compared accuracy across 30 intermediate bianzheng fields to examine its relationship with final syndrome selection (Fig.~\ref{fig:bianzheng_coupling}a). All ten models performed consistently well on summerheat, qi collapse, qi block, exterior/interior, blood cold, qi sinking and fluid retention. For each of these fields, the lowest accuracy across the ten models was at least 90.0\%, while the mean accuracy ranged from 94.7\% to 99.3\%. Qi stagnation, phlegm, kidney and dampness were more difficult, with mean accuracies from 47.3\% to 58.0\%; no model exceeded 70.4\% on any of these fields. Fu-TCM-27B had its largest margins for liver, spleen and internal dampness, three fields on which the reference models were comparatively weak. Its accuracies were 74.6\%, 76.8\% and 75.9\%, exceeding the strongest reference model by 22.8, 18.6 and  19.9 percentage points, respectively. The largest field-level gains for Fu-TCM-27B occurred in several difficult dimensions of zang-fu differentiation and the differentiation of qi, blood and body fluids.

At the model level, mean intermediate bianzheng accuracy was positively correlated with final syndrome-selection accuracy across all ten models (Pearson r = 0.89; Fig.~\ref{fig:bianzheng_coupling}b). Although no detectable association was observed among the eight reference models (Pearson r = $-$0.04, P = 0.93), Fig.~\ref{fig:bianzheng_coupling}b shows that medical domain-specific models tended to lie closer to the overall regression line than general-purpose models. We further restricted the correlation analysis to the four domain-specific models: Fu-TCM-27B, Fu-TCM-9B, HuatuoGPT-3-27B and ShizhenGPT-32B-VL. Within this group, mean intermediate bianzheng accuracy was strongly positively correlated with final syndrome-selection accuracy (Pearson r = 0.996, P = 0.0041; n = 4). This association suggests that TCM-specific training may help models use intermediate diagnostic fields to infer the final syndrome.

To further examine this relationship, we fitted separate case-level logistic-regression models for all ten models. All ten models showed positive associations, with statistical significance reached in nine models (unadjusted \(P<0.05\)). The two Fu-TCM fits are shown in the main text (Fig.~\ref{fig:bianzheng_coupling}c), the eight reference-model fits in Supplementary Figure~3, and the corresponding estimates for all ten models in Supplementary Table 2. For the two Fu-TCM models, the odds ratios per 10-percentage-point increase were 1.88 (95\% confidence interval, 1.35 to 2.63) and 1.52 (1.08 to 2.14), respectively.

We also compared intermediate bianzheng accuracy between cases with correct and incorrect final selections (Fig.~\ref{fig:bianzheng_coupling}d). Across the ten models, mean intermediate accuracy was 2.0 to 7.3 percentage points higher in cases with correct final selections. The bootstrap 95\% confidence intervals excluded zero for nine of the ten models. Both Fu-TCM variants showed this pattern and had the highest overall intermediate accuracies among the evaluated models. Together, these analyses showed positive case-level associations in most evaluated models.

\subsection{Ablation study}

To assess the contribution of the intermediate \textit{bianzheng} reward, we compared domain learning, outcome-only group relative policy optimization (GRPO), and full Bianzheng-Grounded Policy Optimization (BGPO) on the same 311 Fu-bianzheng cases (Fig.~\ref{fig:capability_profiles}d). Full BGPO achieved the highest non-zero intermediate-field accuracy (76.64\%), exceeding domain learning (68.36\%) and outcome-only GRPO (66.70\%). It also attained the highest final syndrome-selection accuracy (73.63\%), compared with 68.17\% and 70.10\%, respectively. Outcome-only GRPO improved final syndrome selection but did not improve intermediate \textit{bianzheng} accuracy, whereas full BGPO improved both measures, supporting the complementary contribution of the intermediate \textit{bianzheng} reward.

\subsection{Performance across textual and visual categories}

\begin{figure}[!htbp]
\centering
\includegraphics[width=\textwidth,height=0.73\textheight,keepaspectratio]{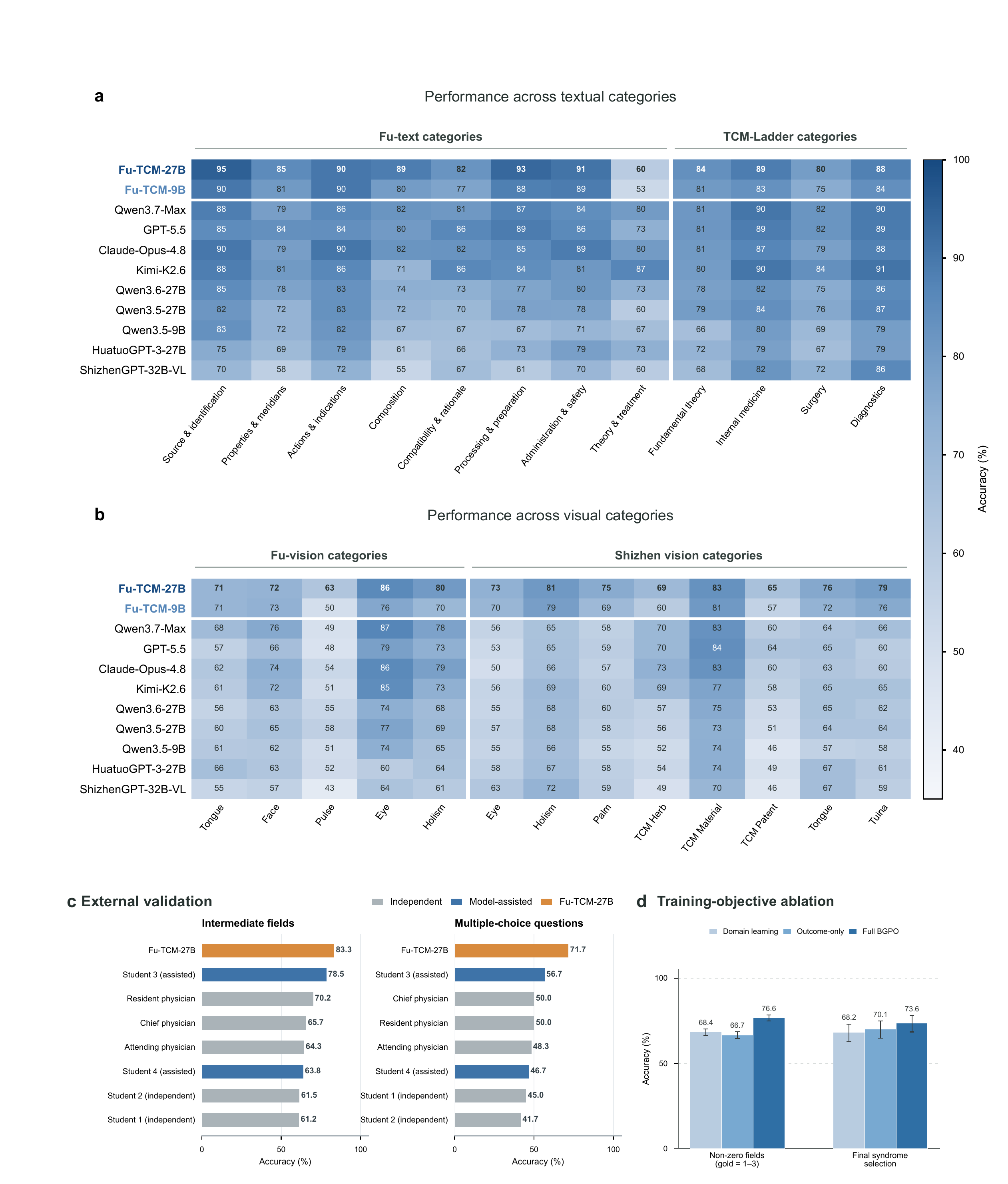}
\caption{Performance across textual and visual categories, external validation and training-objective ablation. a, Accuracy of eleven models across eight Fu-text categories and four TCM-Ladder categories. b, Accuracy of the same models across five Fu-vision categories based on the final item-level classification and eight Shizhen vision categories. Cells show accuracy (\%). c, Individual intermediate-field and multiple-choice accuracy in the external evaluation. Intermediate-field accuracy was assessed on 20 syndrome-differentiation cases, and multiplechoice accuracy was assessed on 60 questions. Grey denotes independent answering, blue denotes model-assisted answering and orange denotes Fu-TCM-27B. Participants are ranked separately within each metric. d, Exact-match accuracy for non-zero intermediate bianzheng fields (reference values 1--3) and final syndrome-choice accuracy after domain learning, outcome-only group relative policy optimization (GRPO) and full Bianzheng-Grounded Policy Optimization (BGPO), evaluated on the same 311 cases. Missing or unparseable intermediate-field predictions were counted as incorrect. Error bars indicate case-level bootstrap 95\% confidence intervals based on 4,000 resamples for non-zero field accuracy and Wilson 95\% confidence intervals for final syndrome-choice accuracy.}
\label{fig:capability_profiles}
\end{figure}

In the text tasks, models generally achieved higher accuracy in actions and indications, internal medicine and diagnostics (Fig.~\ref{fig:capability_profiles}a). This pattern indicates that current models handled questions about actions and indications and diagnostics relatively consistently. Fu-TCM-27B ranked first in five of the eight Fu-text categories and exceeded the strongest non-Fu-TCM model by 5.0 percentage points in source and identification and by 6.4 percentage points in composition. It also achieved the highest accuracy in fundamental theory among the four TCM-Ladder categories. Taken together, these results suggest that reasoning-oriented training may help models integrate TCM textual knowledge more effectively.

In the visual tasks, performance was most consistent for TCM Material, whereas overall performance was lower or more variable for TCM Patent, Pulse, Tongue and Eye (Fig.~\ref{fig:capability_profiles}b). This contrast indicates that direct recognition of clear visual features and TCM diagnostic judgement based on fine-grained visual signs impose different demands on models; the latter requires links to be established between visual information and TCM concepts. Fu-TCM-27B outperformed the strongest non-Fu-TCM model in three of the five Fu-vision categories and ranked first in six of the eight Shizhen vision categories. Its advantages spanned pulse, tongue, eye and palm diagnosis as well as holistic interpretation, indicating a consistent ability to link fine-grained visual signs with TCM diagnostic concepts.

\subsection{External validation}

Following the benchmark evaluation, we conducted an external validation involving TCM physicians. We compared Fu-TCM-27B with physicians answering independently, students answering independently and students answering with model assistance (Fig.~\ref{fig:capability_profiles}c). The participating physicians included a chief physician, an attending physician and a resident physician. The evaluation comprised 20 syndrome-differentiation cases and 60 multiple-choice questions.

Both model-assisted students achieved higher intermediate-field and multiple-choice accuracies than the students who answered independently. Among them, model-assisted Student 3 achieved an intermediate-field accuracy of 78.5\% and a multiple-choice accuracy of 56.7\%, exceeding all three participating physicians on both measures. Compared with the best-performing physician, these accuracies were higher by 8.3 and 6.7 percentage points, respectively; compared with the best-performing independent student, they were higher by 17.0 and 11.7 percentage points, respectively.

Overall, the external validation showed that Fu-TCM-27B can support physicians in syndrome differentiation and answer selection. This assistive function extends the value of the model from standalone task completion to human--AI collaboration and highlights its potential as a tool for supporting physicians in TCM syndrome differentiation and clinical decision-making.

\begin{figure}[!htbp]
\centering
\includegraphics[width=\textwidth,height=0.73\textheight,keepaspectratio]{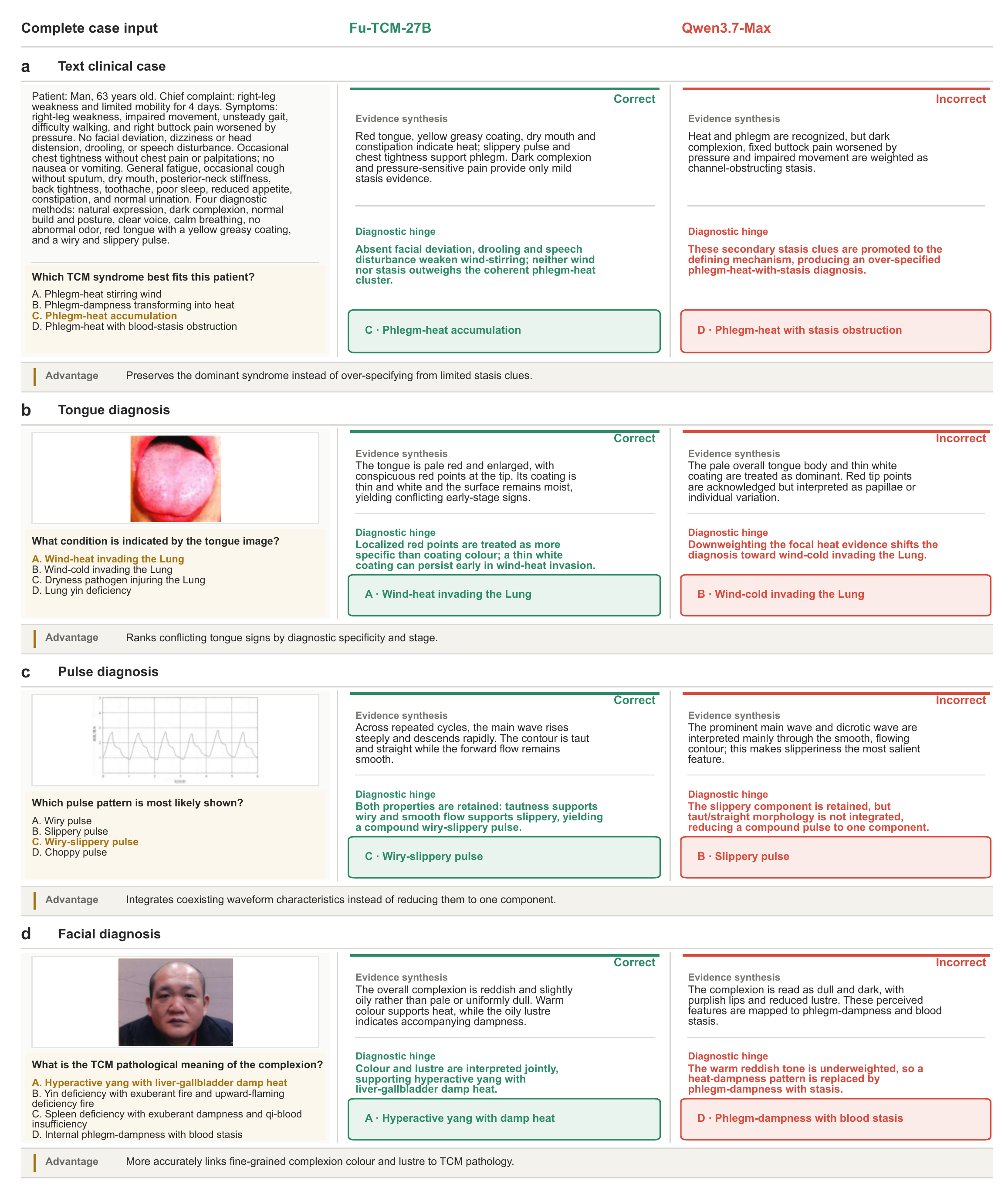}
\caption{Representative case comparisons of TCM diagnostic capabilities. a, Identification of the dominant syndrome from a clinical-text case. b, Recognition of subtle and conflicting tongue features. c, Recognition of compound pulse-waveform characteristics. d, Interpretation of complexion colour, lustre and their pathological meanings. Green and red mark the reasoning and predictions of Fu-TCM-27B and Qwen3.7-Max, respectively; ochre marks the benchmark reference answers.}
\label{fig:case_contrasts}
\end{figure}

\subsection{Case study}

We selected clinical-text, tongue, pulse and facial cases to compare the capabilities of Fu-TCM-27B and Qwen3.7-Max (Fig.~\ref{fig:case_contrasts}).

In the clinical-text case (Fig.~\ref{fig:case_contrasts}a), Fu-TCM-27B integrated the phlegm-heat signs---including a red tongue, yellow greasy coating, dry mouth, constipation, slippery pulse and chest tightness---while treating the dark complexion and pressure-sensitive pain as secondary stasis signs, correctly selecting phlegm-heat accumulation. Qwen3.7-Max overemphasized the localized pain and incorrectly selected phlegm-heat with blood-stasis obstruction, highlighting the advantage of Fu-TCM-27B in distinguishing the dominant syndrome from accompanying manifestations.

In the tongue case (Fig.~\ref{fig:case_contrasts}b), Fu-TCM-27B prioritized the red points at the tongue tip and recognized that a thin white coating could persist during early wind-heat invasion, correctly selecting wind-heat invading the Lung. Qwen3.7-Max relied mainly on the pale-red tongue and white coating and selected wind-cold invading the Lung, showing the stronger capability of Fu-TCM-27B to distinguish subtle and conflicting tongue signs.

In the pulse case (Fig.~\ref{fig:case_contrasts}c), Fu-TCM-27B captured both the taut, straight contour and smooth flow of the waveform and correctly identified a wiry-slippery pulse. Qwen3.7-Max recognized only the smooth-flowing feature and selected a slippery pulse, failing to identify the compound pattern and demonstrating the advantage of Fu-TCM-27B in integrating multiple pulse-waveform characteristics.

In the facial case (Fig.~\ref{fig:case_contrasts}d), Fu-TCM-27B jointly recognized the reddish complexion and oily lustre and correctly identified hyperactive yang with liver--gallbladder damp heat. Qwen3.7-Max perceived the complexion as dark and dull and incorrectly selected internal phlegm-dampness with blood stasis, demonstrating the advantage of Fu-TCM-27B in fine-grained complexion recognition and pathological interpretation.

Together, the four cases demonstrate the broader strengths of Fu-TCM-27B in complex syndrome differentiation, subtle visual understanding, compound pulse recognition and TCM pathological interpretation, complementing the overall quantitative results.

\FloatBarrier
\section{Discussion}

We proposed and validated a training method for TCM large language models grounded in the TCMspecific syndrome-differentiation framework. Strengthening the model's learning of this process improved its performance on TCM text, visual and TCM reasoning tasks. One explanation is that TCM  clinical thinking is itself a reasoning process \cite{zaslawski2003clinicalreasoning,wang2025tcmevalsdt}. Before predicting a syndrome, a model must organize findings from the four examinations into intermediate states such as disease location and nature, zang-fu involvement, and the state of qi, blood and body fluids. This intermediate structure may reduce shortcuts that map isolated symptoms directly to answers and may support more consistent use of TCM knowledge across tasks. Consistent with this interpretation, higher intermediate-field accuracy was associated with more accurate final syndrome selection at the case level in most evaluated models. Together with the ablation results, this finding suggests that explicit training of TCM reasoning structures may complement domain knowledge in improving model performance.

This study has several limitations. First, the structured syndrome-differentiation data used for training came mainly from TCM andrology and cardiology. The disease spectrum, syndrome types and clinical presentations are therefore limited. Generalization to gynaecology, paediatrics, spleen and stomach disorders and other TCM specialties remains to be tested. Second, the reasoning dataset is relatively small and does not fully cover complex syndromes, combined syndromes, long-tail cases or variation among practitioners. This may limit stability in complex clinical settings. The evaluation also relies mainly on fixed syndrome-differentiation fields and multiple-choice tasks. It cannot fully represent open-ended consultation, changes in a patient's condition over time or synthesis across the four examinations in clinical practice.

Training within a TCM reasoning framework improved performance on text, visual and TCM reasoning tasks. Developing TCM-specific models therefore requires more than accumulating domain knowledge; TCM modes of reasoning should also become learnable and assessable training targets. Beyond these performance gains, the method trains the model to organize its outputs according to the TCM reasoning process. By organizing its outputs in this way, the model produces an assessable and traceable record for TCM practitioners. This record provides a basis for reviewing the model's diagnostic rationale and developing trustworthy TCM decision-support systems \cite{who2024lmm,fda2026cds}. Future work should construct larger reasoning datasets that cover more specialties and clinical settings and use them to test the generalizability of this training strategy. This line of work may offer a new design strategy for improving the capabilities of TCM large language models.

\section{Conclusion}

We present Fu-TCM, a multimodal LLM framework that improves TCM models' reasoning capabilities through TCM reasoning training. The framework uses multisource TCM texts, medical images and clinical cases. It combines TCM domain-specific learning with Bianzheng-Grounded Policy Optimization to connect evidence from the four examinations, intermediate fields and final syndrome labels. Fu-TCM-27B achieved the highest accuracy on all six benchmarks and a macro-average of 81.01\%, 7.13 percentage points above the strongest reference model with complete results. On the 311-case Fu-bianzheng benchmark, final syndrome-choice accuracy was 73.63\%, 11.90 percentage points above Qwen3.7-Max, the strongest reference model. For Fu-TCM-27B, strict intermediate-field accuracy was 85.05\%, 9.97 percentage points above GLM-5.2, the strongest reference model on this measure. Intermediate-field accuracy was also positively associated with final syndrome-choice accuracy across the full set of ten evaluated models (Pearson r = 0.89). Positive case-level associations were also observed in most models. The framework retains strong performance across text, vision and TCM reasoning tasks while allowing practitioners to assess and trace how evidence from the four examinations leads to syndrome predictions. This supports review of the model's diagnostic rationale and provides a basis for developing trustworthy TCM decision-support systems.

\FloatBarrier
\section{Methods}

\subsection{Data construction}

We assembled data from classical TCM texts, modern textbooks, image-rich medical books, clinical cases and public datasets. Using DataFlow, we generated text and image--text question--answer pairs from the three types of books \cite{liang2025dataflow}. Text pairs were retained when their answers were supported by the source passages; image--text pairs were retained when the images were correctly aligned with their accompanying captions. Clinical cases were converted into structured records, and public datasets were cleaned and filtered. We confirmed that the processed public data contained no questions from the public benchmarks used for evaluation (Ladder text, Shizhen exam and Shizhen vision). We standardized the resulting records to a common instruction--response format and retained provenance identifiers to record their sources.

For clinical reasoning, each case was represented by evidence from the four examinations, intermediate \textit{bianzheng} fields and a final syndrome set. We constructed the intermediate fields using a clinician-reviewed symptom-to-\textit{bianzheng}-field mapping table. For case \(i\), the input was \(x_i=(e_i^{\mathrm{ins}},e_i^{\mathrm{aus/olf}},e_i^{\mathrm{inq}},e_i^{\mathrm{palp}})\), the structured target was \(z_i=(c_i,o_i)\), and the final syndrome set was \(y_i\subseteq\mathcal S\). The vector \(c_i=(c_i^{\mathrm{EI}},c_i^{\mathrm{CH}},c_i^{\mathrm{DE}},c_i^{\mathrm{YY}})\) encoded the exterior--interior, cold--heat, deficiency--excess and yin--yang categories, with value sets denoted by \(\mathcal C_{\mathrm{EI}}\), \(\mathcal C_{\mathrm{CH}}\), \(\mathcal C_{\mathrm{DE}}\) and \(\mathcal C_{\mathrm{YY}}\), respectively. The exterior--interior field allowed exterior, interior, half-exterior--half-interior, concurrent exterior and interior, or undetermined; each of the other three fields allowed either pole, a mixed state or undetermined. The vector \(o_i\in\{0,1,2,3\}^{26}\) encoded 26 ordinal fields covering five zang-fu dimensions, 15 qi, blood and body-fluid dimensions, and six pathogenic-factor dimensions. Syndrome names followed the \textit{Classification and Codes of Diseases and Patterns of Traditional Chinese Medicine} \cite{gbt15657}. Physicians reviewed a sample of the structured records.

Evaluation used three public benchmarks and three internally constructed benchmarks. The public benchmarks---Ladder text, Shizhen exam and Shizhen vision---were used only for evaluation and were not included in model training. The internally constructed Fu-text and Fu-vision benchmarks comprised 1,000 held-out text question--answer records and 500 held-out image question--answer records, respectively. The images selected for the internally constructed evaluation benchmarks were distinct from those in the training data. For clinical reasoning, 1,867 cases were used for BGPO, and a separate set of 311 cases formed Fu-bianzheng and was excluded from BGPO. Further details of benchmark construction are provided in Supplementary Sections 3.1--3.3.

\subsection{Ethics statement}

The study was approved by the Ethics Committee of the Affiliated People's Hospital of Fujian University of Traditional Chinese Medicine, Fuzhou, China (SQ2025-161-01). Informed consent was obtained from the study participants.

\subsection{Supervised fine-tuning and BGPO}

During TCM domain-specific learning, Qwen3.5-9B and Qwen3.6-27B received full-parameter supervised fine-tuning (SFT) on text and visual question and answer pairs \cite{qwen2026qwen36,zheng2024llamafactory}. For training sample \(n\), input \(x_n\) and serialized target \(q_n\), SFT minimized the target-token-averaged loss
\begin{equation}
\begin{aligned}
M&=\sum_{n,t}m_{nt},\\
\mathcal L_{\mathrm{SFT}}(\theta)
&=-\frac{1}{M}\sum_{n,t}m_{nt}
\log\pi_\theta(q_{nt}\mid x_n,q_{n,<t}).
\end{aligned}
\end{equation}
where \(m_{nt}=1\) for target response tokens and 0 for prompt, padding and non-target visual-placeholder tokens. Thus, only the assistant response contributed to the SFT loss.
Bianzheng-Grounded Policy Optimization (BGPO) sampled \(G=16\) responses per case, with reference syndromes and intermediate fields retained outside the prompt as reward metadata \cite{shao2024grpo,xie2025ladderbase}. Group-relative advantages were
\begin{equation}
A_{ij}=\frac{R_{ij}-\bar R_i}{\sigma(R_i)+\delta},\qquad
\bar R_i=G^{-1}\sum_{j=1}^{G}R_{ij}.
\end{equation}
For token \(t\),
\begin{equation}
\rho_{ijt}(\theta)=
\frac{\pi_\theta(q_{ijt}\mid x_i,q_{ij,<t})}
{\pi_{\theta_{\mathrm{old}}}(q_{ijt}\mid x_i,q_{ij,<t})}.
\end{equation}
The clipped policy objective was \cite{schulman2017ppo}
\begin{equation}
\begin{aligned}
\mathcal L_{\mathrm{BGPO}}
={}&-\mathbb E\!\left[\frac{1}{GT}\sum_{j,t}
\min\!\left(\rho_{ijt}A_{ij},\right.\right.\\
&\left.\left.\qquad
\operatorname{clip}(\rho_{ijt},1-\epsilon_{c},1+\epsilon_{c})A_{ij}
\right)\right]\\
&+\beta D_{\mathrm{KL}}
\!\left(\pi_\theta\Vert\pi_{\mathrm{ref}}\right).
\end{aligned}
\end{equation}
Here, \(\delta=10^{-6}\) is the numerical stabilizer for within-group standardization and \(\epsilon_c=0.2\) is the PPO clipping radius. We set \(\beta=0.001\). Responses used explicit trace tags and a terminal syndrome JSON list; no learned reward or value model was used.

\subsection{Reward function}

The total reward combined format, final syndrome and intermediate \textit{bianzheng}-field scores:
\begin{equation}
R=0.20R_{\mathrm{fmt}}+0.40R_{\mathrm{syn}}+0.40R_{\mathrm{bz}}.
\end{equation}
These weights were selected based on ablation experiments comparing multiple reward-weight configurations. The format reward \(R_{\mathrm{fmt}}\) rewarded the required section order, complete fields, terminal syndrome JSON and limited repetition. The syndrome reward \(R_{\mathrm{syn}}\) equalled 1 for an exact set match; otherwise, the standardized syndrome taxonomy supplied graded credit. For predicted syndrome \(a\) and reference syndrome \(b\), similarity \(s(a,b)\) used node depth \(d(\cdot)\) and their lowest common ancestor \(\operatorname{LCA}(a,b)\) \cite{wupalmer1994}. Best pairwise matches yielded soft precision \(P_s\), soft recall \(Q_s\) and their soft F1:
\begin{equation}
\begin{aligned}
s(a,b)&=\frac{2d(\operatorname{LCA}(a,b))}{d(a)+d(b)},\\
P_s&=\frac{1}{|\hat y|}\sum_{a\in\hat y}\max_{b\in y}s(a,b),\\
Q_s&=\frac{1}{|y|}\sum_{b\in y}\max_{a\in\hat y}s(a,b),\\
R_{\mathrm{syn}}&=\frac{2P_sQ_s}{P_s+Q_s}.
\end{aligned}
\end{equation}
The intermediate reward \(R_{\mathrm{bz}}\) evaluated four categorical eight-principle fields by exact match and 26 ordinal fields (0--3) for zang-fu, qi, blood and body fluids, and pathogenic factors. \(B\) is categorical exact-match accuracy; \(E_w\) is ordinal error weighted by reference intensity; and \(H\) records whether reference-nonzero fields were predicted nonzero. \(N\) combines the error and nonzero-field scores. \(C_c\) and \(C_o\) are the proportions of categorical and ordinal fields successfully parsed, penalizing missing or malformed outputs. Within sums, a satisfied condition contributes 1 and an unsatisfied condition contributes 0:
\begin{equation}
\begin{aligned}
B&=\frac14\sum_{r=1}^{4}\mathbf 1(\hat c_r=c_r),\\
E_w&=\frac{\sum_{k=1}^{26}(1+o_k)|\hat o_k-o_k|}
{\sum_{k=1}^{26}(1+o_k)}.
\end{aligned}
\end{equation}
\begin{equation}
\begin{aligned}
H&=\begin{cases}
\dfrac{\sum_{k=1}^{26}o_k\mathbf 1(\hat o_k>0)}
{\sum_{k=1}^{26}o_k},&\sum_{k=1}^{26}o_k>0,\\[6pt]
1,&\sum_{k=1}^{26}o_k=0,
\end{cases}\\
N&=0.70\left[1-\frac{E_w}{\tau}\right]_+ +0.30H,
\qquad \tau=1.5.
\end{aligned}
\end{equation}
The intensity-weighted error and nonzero-field hit term were designed to discourage indiscriminate zero predictions. When a reference contained no nonzero ordinal field, \(H\) was set to 1 because there was no reference-positive field to recover; any predicted nonzero field remained penalized through \(E_w\). The threshold \(\tau\) was half the maximum fieldwise discrepancy on the 0--3 ordinal scale; weighted errors at or above this threshold therefore received no error-based credit.
\begin{equation}
\begin{aligned}
C_c&=\frac14\sum_{r=1}^{4}\mathbf 1(\hat c_r\ \mathrm{parsed}),\\
C_o&=\frac1{26}\sum_{k=1}^{26}\mathbf 1(\hat o_k\ \mathrm{parsed}),\\
R_{\mathrm{bz}}
&=(0.30B+0.70N)\\
&\quad\times(0.30C_c+0.70C_o).
\end{aligned}
\end{equation}

\subsection{Evaluation and statistical analysis}

Evaluation covered Fu-bianzheng, Fu-text, Fu-vision, Ladder text, Shizhen exam and Shizhen vision \cite{xie2025tcmladder,chen2025shizhen}. Accuracy was the primary metric. For Fu-bianzheng, final syndrome-choice accuracy was the proportion of predictions whose selected option matched the reference option; case-level intermediate accuracy was the proportion of exact matches among 30 fields, with missing fields scored as incorrect.

For Fu-bianzheng, Wilson intervals summarized binomial final-syndrome accuracy. We used 4,000 case-level bootstrap resamples for mean intermediate accuracy and final-correct versus final-incorrect differences, preserving within-case dependence. Across six benchmarks, 20,000 paired item-level resamples compared Fu-TCM-27B with the highest-scoring non-Fu-TCM model. Pearson correlation assessed the model-level association across all ten models and separately across the eight reference models. Logistic regression assessed case-level associations separately within each of the ten models. Odds ratios were reported with 95\% Wald confidence intervals and two-sided, unadjusted \(P\) values. Here, \(Y_i=1\) indicates correct final syndrome selection, \(A_i^{\mathrm{int}}\) is case-level intermediate accuracy, \(\gamma\) its coefficient, and \(\operatorname{OR}_{0.1}\) the odds ratio per 10-percentage-point increase:
\begin{equation}
\begin{aligned}
\operatorname{logit}\Pr(Y_i=1)
&=\alpha+\gamma A_i^{\mathrm{int}},\\
\operatorname{OR}_{0.1}&=\exp(0.1\gamma).
\end{aligned}
\end{equation}

\section{Supplementary Information}

Supplementary information accompanies this manuscript.

\section{Author Contributions}

Jiahang Xie collected, processed, cleaned and curated the data; developed the core codebase and system implementation; performed model training, benchmark experiments and evaluation analyses; and drafted the manuscript. Hongcheng Guo contributed to the core codebase, system implementation and manuscript drafting, conceived and led the project, and supervised the study. Jianfeng Feng provided senior guidance, domain expertise and critical manuscript revision. Jie Huang and Yuanchang Jin contributed to data processing and cleaning. Tong Jin, Pan Sun, Lingfei Wu, Yaping You, Weikang Gong, Wei Cheng and Jia You provided domain expertise and resources and contributed to manuscript revision. All authors discussed the results, reviewed the manuscript and approved the final version.

\section{Competing Interests}

All authors declare no competing interests.

\end{document}